\documentclass[11pt,a4paper]{article}
\usepackage[hyperref]{acl2020}
\usepackage{times}
\usepackage{latexsym}
\usepackage{graphicx}
\usepackage{amsfonts}
\usepackage{danudefs}
\usepackage{booktabs}

\usepackage[disable]{todonotes} 

\makeatletter
\if@todonotes@disabled

\else

\fi
\makeatother

 \newcommand\DB[2][]{\todo[inline, size=\small, caption={2do}, #1]{
\begin{minipage}{\textwidth-4pt}#2\end{minipage}}}

\usepackage{microtype}

\aclfinalcopy 

\title{Contextualised Graph Attention for Improved Relation Extraction}

\author{Angrosh Mandya, Danushka Bollega and Frans Coenen \\
  Department of Computer Science \\
  University of Liverpool, Liverpool, UK \\
  \texttt\{angrosh, danushka.bollegala, coenen\}@liverpool.ac.uk \\
  }

\date{}

\begin{document}
\maketitle
\begin{abstract}
This paper presents a contextualized graph attention network that combines edge features and multiple sub-graphs for improving relation extraction. 
A novel method is proposed to use multiple sub-graphs to learn rich node representations in graph-based networks. 
To this end multiple sub-graphs are obtained from a single dependency tree. 
Two types of edge features are proposed, which are effectively combined with GAT and GCN models to apply for relation extraction. The proposed model achieves state-of-the-art performance on Semeval 2010 Task 8 dataset, achieving an F1-score of 86.3.
\end{abstract}

\section{Introduction}

\begin{figure*}[t]
    \centering
    \includegraphics[width=0.75\textwidth]{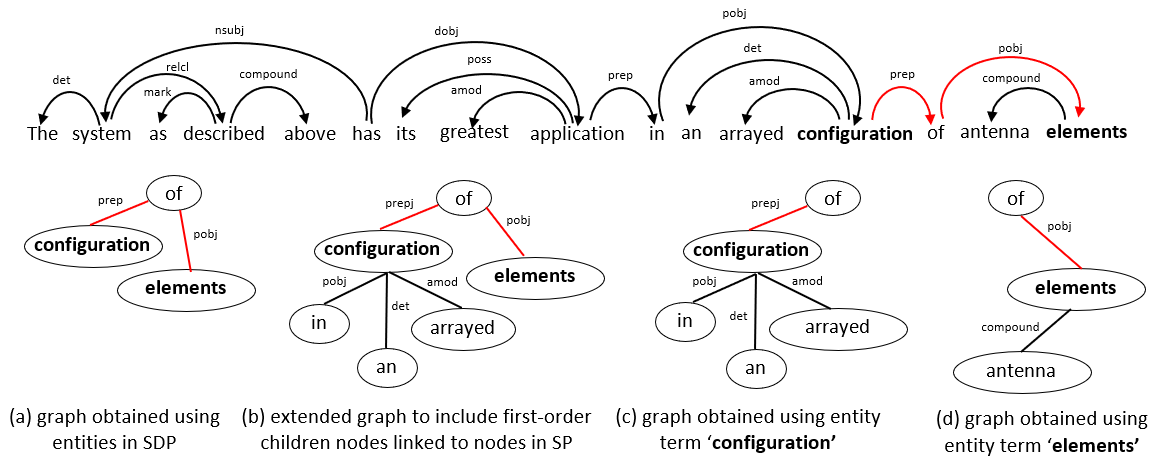}
    \caption{(a). Dependency graph for the example sentence; (b) to (e) Various sub-graphs obtained from the dependency graph for the sentence.}
    \label{fig:graph_example}
\end{figure*}

Recently, Graph Convolution Networks (GCNs) have shown promising results for relation extraction \cite{schlichtkrull2018modeling,zhang2018graph,guo2019attention,fu2019graphrel}. 
GCNs generalises the convolution operation from traditional data such as images and grids to graphical data and generates vertex representations by aggregating features from neighbouring vertices and as well as the features associated with those vertices.
\DB{Better to stick to vertex/edge (terminology from graph theory) or node/link (terminology from network theory) and not mix those. Also lets stick to UK spelling as we are in UK now}
In the context of relation extraction, the graphical structure for sentences is obtained using methods such as:
(a) dependency trees \cite{zhang2018graph}; (b) adjacent edges across consecutive words \cite{peng2017cross}; and (c) co-reference and discourse relations between sentences \cite{peng2017cross}. 
In the case of using dependency tree structures, the words in the sentence serve as vertices in the graph and the dependency relations between words provide the edges between vertices in the graph. 
Further, graphs of different sizes can be derived using the dependency parse tree. 
For example, for the sentence shown in Figure \ref{fig:graph_example}, a small-sized graph containing three vertices (Figure \ref{fig:graph_example}(a)) can be obtained by using vertices in the shortest dependency path (SDP) between entities \textbf{``configuration''} and \textbf{``elements''}. 
\DB{Avoid single quotes in scientific writing as they are informal}
The same graph can be extended by including first-order child vertices connected to the vertices in the SDP (shown in Figure \ref{fig:graph_example}(b)).

Although GCNs are useful for relation extraction, providing the appropriate graph structure with important vertices and edges is vital to achieve optimum performance. 
While using small-sized graphs would eliminate useful information from graphs, large-sized graphs can add more noise, resulting in difficulties for the network to learn useful vertex representations. 
For example, while the Contextualised Graph Convolution Network (C-GCN) \cite{zhang2018graph} achieves a higher performance with graphs using first-order child vertices connected to vertices in SDP (such as the ones shown in Figure \ref{fig:graph_example}(b)), it's performance significantly drops when the graph is limited to the vertices in SDP (Figure \ref{fig:graph_example}(a)) or when a higher number of child vertices (second order and above) are included in the graph. 
\DB{This is the first mention of C-GCN. Give the full form. Contextualised Graph Convolutional Networks?}
Moreover, GCNs are known to struggle on large-sized graphs derived from real-world datasets such as protein structures \cite{borgwardt2005protein} and HIV infected patients' data \cite{ragin2012structural,zhang2016identifying}.
The noisy nature of graphs involving complex vertex features and edges complicates the learning for GCNs. 
To address the problem of dealing with large and noisier graphs, \newcite{lee2018graph}  proposed graph attention model (GAM) to learn to discriminate patterns confined to specific regions in the large graph. 
\DB{Use the newcite command to get the desired Author-Year format in ACL bst}

With a focus to reduce complexity in learning from large graphs, we propose to use multiple sub-graphs as opposed to using a single graph with graphical networks. 
Specifically, we derive multiple sub-graphs from a single dependency tree for the task of relation extraction. 
We propose a novel method to obtain sub-graphs using vertices corresponding to the target entities in the sentence as shown in Figures \ref{fig:graph_example}(c) and \ref{fig:graph_example}(d). 
Thus, instead of using a larger graph such as the one shown in Figure \ref{fig:graph_example}(b), we propose to use multiple sub-graphs (shown in Figures \ref{fig:graph_example}(a), (c) and (d)) to jointly learn for relation extraction. 
Using such segregated structures would facilitate focusing on specific regions, useful for learning richer representations, particularly for the vertices corresponding to the target entities.

Further, more recently, graph attention networks (GATs) \cite{velivckovic2017graph} are shown to achieve superior performance for vertex classification in graph structured data. 
In contrast to GCNs that aggregate neighbouring vertices as features to generate vertex representations, GATs attend over neighbourhood vertex features to compute weights for learning vertex representations. 
In this paper, we propose to use GATs for relation extraction. 
Although GATs~\cite{velivckovic2017graph} consider the importance of neighbouring vertices for deriving vertex representations, GATs do not consider the edge features for computing attention weights. 
Recently \newcite{gong2018exploiting} showed that by combining edge features with GATs improves vertex classification. 
In the context of relation extraction, edge features can provide useful clues to identify relations across entities. 
For example, the information of vertices connected to different entity types or the dependency relation between vertices can serve as useful features when computing the salience of neighbouring vertices. 
Given this aspect, we proposes a contextualised GAT that combines edge features for relation extraction. 
The key contributions of this paper are:
\begin{itemize}
    \item Propose a contextualized graph attention network that combines edge features from multiple sub-graphs for relation extraction.
    \item Present a novel method to derive sub-graphs using dependency parse and entity positions.
    \item Combining dependency relations and entity type features with GATs and GCNs.
    \item Conduct an empirical comparison between graphical networks (GCNs and GAT) using single-graph vs. multiple sub-graphs.
\end{itemize}
Our proposed method achieves the state-of-the-art (SoTA) performance for relation extraction on the Semeval 2010 Task 8 relation extraction benchmark dataset.





\section{Related Work}

Various graph-based neural networks are shown to improve relation extraction. 
\newcite{xu2015classifying} applied LSTMs over SDP between the target entities to generalise the concept of dependency path kernels. \newcite{liu2015dependency} used RNNs to model sub-trees in the dependency graph and a CNN to capture salient features from the SDP. \newcite{miwa2016end} used Tree-LSTMs \cite{tai2015improved} and BiLSTMs on dependency tree structures to jointly model entity and relation extraction. 
\newcite{zhang2018graph} proposed C-GCNs for relation and proposed a pruning strategy to selectively include vertices in the graph structure. \newcite{guo2019attention} presented Contextual-Attention Guided Graph Convolutional Networks (C-AGGCN) to selectively attend to important parts in the dependency graph to learn rich node representation in the graphical network. 
\DB{First mention of C-AGGCN. Give the full form name within brackets when you introduce an acronym for the first time.}
On the other hand, the key focus of this paper is to combine multiple sub-graphs for relation extraction as opposed to learning from a single graph as in the case of C-GCN and C-AGGCN. 
For this purpose, we modify GAT \cite{velivckovic2017graph} and incorporate novel edge features for relation extraction. 
To the best knowledge of the authors, this is the first study that proposes a contextualised graph attention network with edge features, to learn from multiple sub-graphs for improved relation extraction.
\DB{Although not limited to this paper, I prefer writing each sentence in a separate line in \LaTeX source.That makes commenting out and editing easier, and also shows the logical structure of the text more clearly.}

\section{Contextualised Graph Attention with Edge Features over Multiple Graphs} \label{sec_3_c_gcn_mg}


\subsection{Problem Formulation}

\begin{figure}[t]
    \centering
    \includegraphics[width=0.4\textwidth]{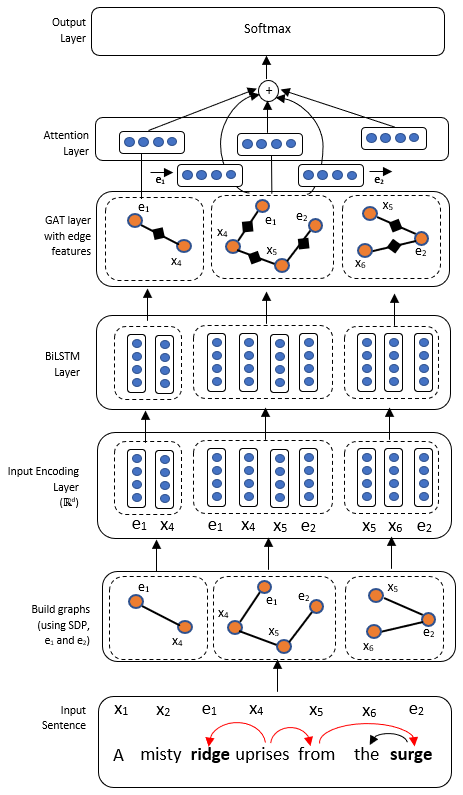}
    \caption{Architecture of the contextualised graph attention network incorporating edge features and conduct relation extraction using multiple sub-graphs.}
    \label{fig:gcn_architecture}
\end{figure}

Let $ \mathcal{S} = [s_1,...,s_k]$ denote a set of sentences, where each sentence $s_i = [x_1,...x_n]$ is a set of tokens where $x_i$ is the $i$-th token. Further, each $s_i$ also consists of two target entities $e_1$ and $e_2$ between which a semantic relation $r$ exists, selected from a pre-defined set of relations $\mathcal{R}$. 
Thus, given $s_i, e_1$ and $e_2$, the relation extraction task is to identify the relation $r$ that holds between entities $e_1$ and $e_2$. 
The set $\mathcal{R}$ also contains the label ``no relation'', which is predicted when there exists no relation between the two entities. 
In this study, we formulate the relation extraction task as a \emph{graph classification} task. 
For this purpose, each $s_i$ is represented as a set of sub-graphs $g_i^k$, where $k$ is the number of sub-graphs. 
Numerous operations are performed on the individual sub-graphs using a contextualised graph attention that combines edge features to learn rich vertex representations for relation extraction. 
The architecture of the proposed model is shown in Figure \ref{fig:gcn_architecture} and is further explained below.


\subsection{Modelling Sentences using Sub-Graphs}

The sentence $s_i$ and entity mentions $e_1$ and $e_2$ are provided as the input to the proposed model as seen in Figure \ref{fig:gcn_architecture}. 
\DB{We already defined token sequence in the previous section. So I dropped it.}
In the first step, multiple sub-graphs for a sentence is created using the dependency parse of the sentence. 
Specifically, three sub-graphs are obtained using SDP and positions of the target entities $e_1$ and $e_2$. 
For instance, for the example sentence in Figure \ref{fig:gcn_architecture}, the following three sub-graphs are obtained: 
(a) graph comprising vertices (``ridges'', ``uprises'', ``from'', ``surge'') in SDP; 
(b) graph comprising vertices (``ridges'', ``uprises'') connected to the entity $e_1$; 
(c) graph comprising vertices (``surge'', ``from'', ``the'') connected to the entity $e_2$. 
Although dependency parse defines the direction between related words, it is ignored to obtain undirected sub-graphs for the sentence as seen in Figure \ref{fig:gcn_architecture}. 
We separately create an adjacency matrix $(\mathbf{A}^k)$ to preserve the ordering of vertices.
\DB{How is this matrix created?}
\begin{align}
    s_i = {g_i^k},\quad \mbox{ where} \: k=1,2,3
\end{align}
\DB{Use the align enviornment instead of equation and eqnarray environments for math.}

\subsection{Input encoding layer}

Each sub-graph $g_i^k$ consists of a sequential set of tokens $[x_1^k, \ldots ,x_n^k ]$ (although some words could be missing), which is encoded into a fixed length vector using (a) contextual; (b) part-of-speech; (c) dependency; (d) named entity type; and (e) word type embeddings. BERT \cite{devlin2018bert} is used to obtain the contextual embeddings. 
BERT tokenises each token $x_i$ into $s$ Byte-Pair Encoding~\cite[BPE;][]{Sennrich:2016}  tokens ($x_i = \{b_1, b_2, \ldots ,b_s\}$) and generates $L$ hidden states for each BPE token, $\mathbf{h}_t^l, 1 \leq l \leq L, 1 \leq t \leq s $. The contextual embedding for each token is obtained by summing the last four layers of the BERT model.\footnote{Our preliminary experiments showed that using the last four layers resulted in the best performance.} 
Thus, the token encoding, $\mathbf{BERT}_{x_i}$ of the token $x_i$ is given by \eqref{eq:BERT}.
\begin{align}
    \label{eq:BERT}
    \mathbf{BERT}_{x_i} = \sum_{l = L-4}^L \frac{\sum_{t=1}^s \mathbf{h}_t^l}{s}. 
\end{align}
Additionally, for each token $x_i$, a $p$-dimensional feature vector is created to represent 
(a) Part-of-Speech (POS) tags ($f_{x_i}^{pos}$); 
(b) dependency relations ($f_{x_i}^{dep}$); 
and (c) and named entity types ($f_{x_i}^{net}$). 
Moreover, a $q$-dimensional feature vector is created to indicate whether the token $x_i$ is an entity mention or not ($f_{x_i}^{wt}$). 
Thus, the final input vector for each token $x_i$ is given by \eqref{eq:tokrep}.
\begin{align}
    \label{eq:tokrep}
    \tilde{x}_i^k = [\mathbf{BERT}_{x_i^k};f_{x_i^k}^{pos}; f_{x_i^k}^{dep}; f_{x_i^k}^{net}; f_{x_i^k}^{wt}] 
\end{align}
$f_{x_i}^{pos}$, $f_{x_i}^{dep}$, $f_{x_i}^{net}$ and $f_{x_i}^{wt}$ are randomly initialised and updated during training, whereas $\mathbf{BERT}_{x_i}$ is computed according to \eqref{eq:BERT} using a pre-trained BERT model.
\DB{I added the last line as there was no mention as to how we get these additional embeddings. Did I say that correctly?}


\subsection{Contextual BiLSTM layer}


To further fine-tune the input embeddings, the encoded series of input vectors $[\tilde{x}_1^k, \ldots ,\tilde{x}_n^k ]$ in each graph is provided as the input to a contextualised BiLSTM layer to produce a $d^l$-dimensional hidden state vector for each input token in both forward and backward directions as given in \eqref{eq:BiLSTM}.
\begin{align}
    \label{eq:BiLSTM}
     \mathbf{\vec{h}}_1^k, \ldots ,\mathbf{\vec{h}}_n^k = \mathbf{BiLSTM}(\tilde{x}_1^k, \ldots ,\tilde{x}_n^k)
\end{align}
Here, $\mathbf{h}_i \in \mathbb{R}^{2d^l}$, where $d^l$ is the dimension of hidden state of the LSTM and $\mathbf{h}_i$ is the hidden state vector of BiLSTM at time-step $i$, considering both forward and backward directions.
The BiLSTM layer is jointly trained along with the rest of the model.


\subsection{Graph Attention Layer with Edge Weights}
\label{sec_gat}

\newcite{velivckovic2017graph} proposed GATs to assign larger weights to important vertices in the graph by computing attention weights across neighbouring vertices.
\DB{We defined the acronym of GAT already in the intro. Do not define the same acronym multiple times in the same paper}
We modify the GAT to include the following edge features to compute attention weights for deriving vertex features in each sub-graph.

\subsubsection{Edge Features} \label{sec_edge_features}

In addition to neighbouring vertices, edge features are useful for learning rich vertex representations, particularly in the context of relation extraction. 
For example, the information of vertices in the graph that are connected to entity mentions are helpful in providing higher weights for vertices connected to entity mentions to facilitate accurate relation extraction. 
Similarly, dependency relations between vertices can serve as useful features to improve relation extraction. 
Given the usefulness of edge features, we evaluate the following two types of edge features in combination with GAT.
\paragraph{Dependency relations based edge features (\textsc{dref}):}
The \textsc{dref} features are derived based on the frequency of the dependency relations that exist between vertices in the training corpus as follows. Let  $\mathcal{D}$ and $\mathcal{P}$ be the sets of respectively all dependency relations across vertices and Part-Of-Speech (POS) tags of vertices observed in the training corpus. 
The edge weight for a given dependency relation $r^d \in \mathcal{D}$ between vertices $i$ and $j$ with respectively POS tags $\lambda$ and $\mu \in \mathcal{P}$ ($\lambda=\mu$, where vertices $i$ and $j$ have the same POS tag) is defined as the ratio of total number of times the triple $(\lambda, \mu, r^d)$ is encountered in the corpus to the total number of triples across all vertices (POS tags) with different dependency relations in the corpus. 
Each of these edge weights is assigned an $d^e$-dimensional random feature vector and subsequently updated along with the GAT in order to compute attention weights for deriving vertex representations.
\DB{I replaced $p$ and $q$ as these were used previously for the dimensions of the feature vectors.}
\paragraph{Connection type edge features (\textsc{ctef}):}
The \textsc{ctef} features are used to identify whether a given node is connected to an entity term or a non-entity term in the graph. 
Because GCNs and GATs operate on undirected graphs, such information is not available to the network and providing information about vertices connected to entities can improve performance. 
Thus, a $d^{e}$-dimensional feature vector of ones is defined for edges where the out-going node is an entity mention, or a $d^{e}$-dimensional feature vector of zeros is included otherwise, and combined along with the GAT for computing vertex representations.

\subsubsection{Graph Attention Operation}

The output from the contextual BiLSTM layer combined with edge features (defined above) is provided as input to the GAT layer to produce a new set of $m$-dimensional vertex representations.
\DB{We used $f$ above for feature vectors. Can we use a different symbol? $m$ for example? }
Following \newcite{velivckovic2017graph}, to derive higher-level features from the input features, a shared linear transformation, parametrised by a \emph{weight martix}, $\mat{W} \in \R^{{2d^l} \times m} $, is applied to every vertex. 
In order to perform \emph{self-attention} on vertices, a shared attention mechanism $\alpha : \mathbb{R}^{m+{d^e}} \times \mathbb{R}^{m+{d^e}} \rightarrow \mathbb{R} $ is used to compute the \emph{attention coefficients}, $e_{ij}$, that indicate the importance of vertex $j$'s features to vertex $i$ as given in \eqref{eq:attention}.
\begin{align}
    \label{eq:attention}
     e_{ij} = \alpha(\mathbf{W}\vec{h}_i, \mathbf{W}\vec{h}_j, \vec{e}_{ij}^f) 
\end{align}
Here, $\vec{e}_{ij}^f$ is $d^e$-dimension edge feature vector connecting vertex $j$ to vertex $i$, obtained as described earlier. The graph structure is injected into the attention mechanism by computing $e_{ij}$ for vertices $j \in \mathcal{N}_i$, where $\mathcal{N}_i$ is some neighbourhood of the vertex $i$ in the graph. 
The coefficients are normalised across all choices of $j$ using the softmax function to make them comparable across different vertices as given by \eqref{eq:softmax}.
\begin{align}
    \label{eq:softmax}
    \alpha_{ij} = \mathrm{softmax}_j(e_{ij}) = \frac{\exp(e_{ij})}{\sum_{k \in \mathcal{N}_i} \exp (e_{ik})} 
\end{align}
The attention mechanism, $\alpha$, is a single-layer feed-forward network, parametrised by a weight vector $\vec{\mathbf{\alpha}} \in \mathbb{R}^{m+{d^e}} $. 
Applying a non-linearity function, the attention coefficients is given by \eqref{eq:nonlin}.

{\small
\begin{align}
    \label{eq:nonlin}
    \alpha_{ij} = \frac{\exp(\mbox{L}(\vec{\alpha}\T [\mathbf{W}\vec{h}_i \oplus \mathbf{W}\vec{h}_j \oplus \vec{e}_{ij}^f]))}{\sum_{k \in \mathcal{N}_i} \exp(\mbox{L}(\vec{\alpha}\T [\mathbf{W}\vec{h}_i \oplus \mathbf{W}\vec{h}_j \oplus \vec{e}_{ij}^f]))}
\end{align}
}%
Here, $\mbox{L}$ is the $\mbox{LeakyReLU}$ non-linearity function with negative slope $\alpha=0.2$, $\T$ is transposition and $\oplus$ denotes vector concatenation.
\DB{|| is used in vector norm. $\oplus$ is better}
The normalised attention coefficients considering adjacent vertex and edge features are linearly combined with corresponding vertex features to obtain the final output representation for each vertex as given by \eqref{eq:vertrep}.

{\small
\begin{align} 
    \label{eq:vertrep}
    \vec{h}_i^l = \sigma \left( \sum_{j \in \mathcal{N}} \alpha_{ij} \mathbf{W} \vec{h_j} \right)
\end{align}
}%

\newcite{velivckovic2017graph} found that extending the attention mechanism to \emph{multi-head attention} was further beneficial for vertex classification. 
Consequently, we use multi-head attention on the combination of vertex and edge features. 
Specifically, the transformation given by \eqref{eq:vertrep} is executed $K$ independent times and the resulting features are concatenated to obtain the vertex representation in \eqref{eq:multhead} for individual vertices.

{\small
\begin{align}
    \label{eq:multhead}
    \vec{h}_i^l = \bigoplus_{k=1}^K \sigma \left( \sum_{j \in \mathcal{N}} \alpha_{ij}^k \mathbf{W^k} \vec{h_j} \right)
\end{align}
}%

\subsection{Attention layer}

The output of the graph attention layer combined with edge features is the vertex-level output $\mathbf{Z} \in \mathbb{R}^{n \times m}$, where $n$ is the number of nodes and $m$ is the dimensionality of the output features. 
Intuitively, the feature representation of each vertex is an aggregation of information from the connecting neighbouring vertices and edge features in the graph. 
In order to derive the final representation to be used for relation classification, a final attention layer is used to determine each vertex's contribution and derive a fixed-length feature vector for the graph $g_i$. 
The attention mechanism in the final attention layer assigns a weight $\alpha^\prime_i$ to each vertex annotation $\vec{h}_i^l$. 
A fixed-length representation $v_{g_i} \in \mathbb{R}^{d^g} $ is computed for the graph $g_i$, as the weighted-sum of all vertex annotations as given by \eqref{eq:final-att}.

{\small
\begin{align}
    \label{eq:final-att}
      v_{g_i} &= \sum_{i=1}^N \alpha^\prime_i \vec{h}_i^l, \ \ \  v_{g_i} \in \mathbb{R}^{d^g} \\
      \text{where,   } u_i &= \tanh(\mat{W}^\prime \vec{h}_i^l) \\
    \alpha^\prime_i &= \frac{\exp(u_i)}{\sum_{t=1}^T \exp(u_t)}, \mbox{   } \sum_{i=1}^T \alpha_i = 1
\end{align}
}%
Here, $\alpha^\prime_i, \mat{W}^\prime$ indicate the parameters of the attention layer on top of GAT. The final representation for the sentence is obtained by summing all the three vectors obtained for the three graphs along with the corresponding hidden state vectors of entity mentions $e_1$ and $e_2$ obtained at the GAT layer as given by \eqref{eq:sentrep}.

{\small
\begin{align}
    \label{eq:sentrep}
    v = \sum_{i=1}^2 \vec{h}_{e_i}^{l} +  \sum_{k=1}^3 v_{g_i^k}, \mbox{  } h_{e_i}^l \in \mathbb{R}^{d^g}, \text{           } v_{g_i^k} \in \mathbb{R}^{d^g}
\end{align}
}
Here, $\vec{h}_{e_i}^{l}$ is the hidden state vector of entity mentions $e_i$ at layer $l$ of GAT.

\subsection{Output Layer}

The final feature vector from attention layer $v \in  \mathbb{R}^{d^g}$ is provided as input to a fully connected softmax layer to obtain a probability distribution over relation types. 
The cross-entropy loss for label prediction is given by \eqref{eq:CE}.
\begin{align}
    \label{eq:CE}
    J(\theta) = \sum_{i=1}^r \log p(r_q | v, \theta)
\end{align}
Here, $r$ is the total number of relation types and $\theta$ are the parameters of the model. 
During inference, the test instances are represented as graphs and fed to the trained classifier to predict the corresponding relation type.

\section{Experiments}


\subsection{Dataset}

We evaluate the proposed method on the SemEval-2010 Task 8 dataset (SemEval), which contains 10,717 sentences (8,000 train and 2,717 test), with each sentence marked with two nominals ($e_1$ and $e_2$) and labelled with a relation $r$ from a set of 9 different relation types and an artificial relation ``Other''. 
The task is to predict the relation between the nominals considering the directionality. 
Following prior work, we report the official macro F1-Score excluding the `Other' relation as the evaluation measure.

\subsection{Implementation Details}

The proposed model is implemented using PyTorch\footnote{\url{https://pytorch.org/}}.
Spacy \cite{spacy2} is used to obtain dependency trees, POS tags, named entity types, and dependency relations. PyTorch Geometric \cite[PyG;][]{Fey/Lenssen/2019}, is used to implement GCN and GAT with combined edge features. The hyperparameters of the model were tuned using a development set obtained by randomly selecting 10\% of the training set. 
\DB{It is obvious you tested on the test set. I removed test set from that sentence because that sentence is about hyperparameter tuning}
The model was trained for 200 iterations following mini-batch gradient descent (SGD) with a batch size of 50. Word embeddings were initialised using 768-dimensional contextual BERT embeddings. The dimensions for embeddings for part-of-speech (POS), named entity tags, dependency tags was set to 40 and were initialised randomly. The dimensions for word-type embeddings was set to 10. The dimensions of hidden state vector in the LSTM, GCN, GAT and attention layer was set to 256.

\subsection{Evaluated Models}

Given the two types of edge features used in this study (described in section \ref{sec_edge_features}), the contextualised graph attention network over multiple graphs (\textsc{c+gat+mg}) using different sets of edge features: \textsc{c+gat+mg+dref}; \textsc{c+gat+mg+ctef}; \textsc{c+gat+mg+dref+ctef} are evaluated against various baseline models as listed in Table \ref{table:baselines}.

\begin{table}[t]
\begin{center}
{\small
\begin{tabular}{p{7.0cm}}
\hline
(1) \textsc{c+gat+mg} with out edge features; \\
\hline
(2) \textsc{gat} using \textit{multiple} graphs with different edge features: \textsc{gat+mg}; \textsc{gat+mg+ctef}; \textsc{gat+mg+dref}; \textsc{gat+mg+ctef+dref}; \\
\hline
(3) \textsc{gat} using \textit{single} graph with different edge features: \textsc{gat+sg}; \textsc{gat+sg+ctef}; \textsc{gat+sg+dref}; \textsc{gat+sg+ctef+dref}; \\
\hline
(4) contextualized \textsc{gcn} using multiple graphs with various edge features: \textsc{c+gcn+mg}; \textsc{c+gcn+mg+ctef}; \textsc{c+gcn+mg+dref}; \textsc{c+gcn+mg+ctef+dref}; \\
\hline
(5) contextualized \textsc{gcn} using single graph with different edge features: \textsc{c+gcn+sg}; \textsc{c+gcn+sg+ctef}; \textsc{c+gcn+sg+dref}; \textsc{c+gcn+sg+ctef+dref}; \\
\hline
(6) \textsc{gcn} using multiple graphs with various edge features: \textsc{gcn+mg}; \textsc{gcn+mg+ctef}; \textsc{gcn+mg+dref}; \textsc{gcn+mg+ctef+dref}; \\
\hline
(7) \textsc{gcn} using single graph with various edge features: \textsc{gcn+sg}; \textsc{gcn+sg+ctef}; \textsc{gcn+sg+dref}; \textsc{gcn+sg+ctef+dref}.\\
\hline
\end{tabular}
}
\end{center}
\caption{\label{table:baselines} Various baselines evaluated in the study}
\vspace{-5mm}
\end{table}


\DB{I removed the subsection to get space and avoid subsubsections}

\subsection{Influence of Edge Features}

The performance of various models using different sets of vertex features is shown in Table \ref{table:single_graph_vs_multi_graph}. The \textsc{c+gat+mg+dref} model using dependency relation-based edge features achieves the best F1-score of 86.30. 
The \textsc{c+gat+mg+dref} model scores both in terms of higher precision (86.03) and recall (86.66) to achieve a higher F1-score, indicating the usefulness of the proposed model. 
The GCN models using \textsc{dref} and \textsc{ctef} edge features also report comparatively higher F1-scores of 85.82 and 85.91, respectively. 
The higher performance of \textsc{c+gat+dref} model mainly due to its higher recall (87.56). 
However, \textsc{c+gat+ctef} scores a higher precision (87.24) using \textsc{ctef} features. 

As clearly evident from Table \ref{table:single_graph_vs_multi_graph}, combining edge features along with vertex features generally helps to obtain superior performance in comparison to models that rely only on vertex features. 
For example, the performance of \textsc{c+gat+mg} model using both vertex features and edge features (\textsc{ctef} and \textsc{dref}) is higher than that of \textsc{c+gat+mg}, which considers only vertex features. 
A similar result is observed across other multiple graph-based models such as \textsc{gat+mg}, \textsc{c+gcn+mg} and \textsc{gcn+mg}. 
Although a higher performance is achieved using \textsc{dref} and \textsc{ctef} edge features individually, combining \textsc{dref} and \textsc{ctef} does not help in further improving the performance.

We see that using a contextual layer to provide vertex features for GCN and GAT layers is useful for obtaining better performance. As seen in Table \ref{table:single_graph_vs_multi_graph}, both GCN and GAT models using a contextual layer achieve higher F1-scores over their non-contextual counterparts for both single and multiple sub-graphs. 
The performance of GAT-based models is comparatively higher than GCN-based models for relation extraction. The ability of GATs to attend to neighbouring vertices and edge features when computing vertex representations is more useful than simply considering the structural information as in the case of GCNs to achieve higher performance.

\begin{table}[t]
\begin{center}
{\small
\begin{tabular}{l c c  c}
\toprule
Model & P & R & F \\
\multicolumn{4}{c}{\textsc{single graph (sg)}} \\
\midrule
\textsc{gcn+sg} & 81.82 & 79.44 & 80.50 \\
\textsc{gcn+sg+ctef} & 80.25 & 80.57 & 80.28 \\
\textsc{gcn+sg+dref} & 81.49 & 82.15 & 81.74 \\
\textsc{gcn+sg+ctef+dref} & 81.32 & 80.06 & 80.50   \\
\midrule
\textsc{c+gcn+sg} & 83.16 & 84.62 & 83.84 \\
\textsc{c+gcn+sg+ctef} & 82.80 & 84.28 & 83.44 \\
\textsc{c+gcn+sg+dref} & 82.91 & 84.02 & 83.39 \\
\textsc{c+gcn+sg+ctef+dref} & 82.66 & 83.92 & 83.16 \\
\midrule
\textsc{gat+sg} & 81.34 & 80.40 & 80.80 \\
\textsc{gat+sg+ctef} & 83.85 & 78.27 & 80.78 \\
\textsc{gat+sg+dref} & 81.92 & 81.44 & 81.63 \\
\textsc{gat+sg+ctef+dref} & 81.49 & 80.98 & 81.13 \\
\midrule
\textsc{c+gat+sg} & 82.18 & 85.04 & 83.52 \\
\textsc{c+gat+sg+ctef} & 83.21 & 82.80 &  82.87 \\
\textsc{c+gat+sg+dref} & 82.28 & 83.72 & 82.89 \\
\textsc{c+gat+sg+ctef+dref} & 83.11 & 82.98 & 82.94 \\
\midrule

\multicolumn{4}{c}{\textsc{multiple graphs (mg)}} \\
\hline
\textsc{gcn+mg} & 83.18 & 85.89 & 84.40 \\
\textsc{gcn+mg+ctef} & 87.97 & 82.49 & 84.99 \\
\textsc{gcn+mg+dref} & 84.39 & 84.76 & 84.52 \\
\textsc{gcn+mg+ctef+dref} & 85.57 & 84.04 & 84.74 \\
\midrule
\textsc{c+gcn+mg} & 86.28 & 83.63 & 84.83 \\
\textsc{c+gcn+mg+ctef} & \textbf{87.24} & 84.65 & 85.82 \\
\textsc{c+gcn+mg+dref} & 84.43 & \textbf{87.56} &  85.91 \\
\textsc{c+gcn+mg+ctef+dref} & 83.53 & 88.08 & 85.40 \\
\midrule
\textsc{gat+mg} & 86.08 & 83.76 & 84.80 \\
\textsc{gat+mg+ctef} & 84.98 & 84.33 & 84.61 \\
\textsc{gat+mg+dref} & 85.37 & 85.06 & 85.16 \\
\textsc{gat+mg+ctef+dref} & 85.37 & 85.06 & 85.16 \\
\midrule
\textsc{c+gat+mg} & 86.84 & 83.64 & 85.12 \\
\textsc{c+gat+mg+ctef} & 86.60 & 84.85 & 85.62 \\
\textsc{c+gat+mg+dref} & 86.03 & 86.66 & \textbf{86.30} \\
\textsc{c+gat+mg+ctef+dref} & 84.94 & 85.53 & 85.16 \\
\bottomrule
\end{tabular}
}
\end{center}
\caption{\label{table:single_graph_vs_multi_graph} Performance of various models on SemEval test set. P: Precision; R: Recall; F: F1-score}
\vspace{-3mm}
\end{table}
\DB{Use booktabs for nicer tables}

\subsection{Using Single vs. Multiple Graphs}

It is evident from Table \ref{table:single_graph_vs_multi_graph} that by using multiple sub-graphs instead of a single graph is beneficial for relation extraction. 
The performance of \textsc{gcn, c+gcn, gat} and \textsc{c+gat} models using \textit{multiple} graphs scores significantly higher than its counterparts using only a \textit{single} graph. 
The improvement is not limited to models that use edge features but holds true for models that do not use edge features as well. 
For example, results of  \textsc{c+gat+mg} (F1-score of 85.12), which uses multiple sub-graphs (without edge features) is higher than \textsc{c+gat+sg} (F1-score of 83.52), which uses a single graph. 
The same results can be seen across other models: \textsc{gat+mg (84.80) vs. gat+sg (80.80)}; \textsc{c+gcn+mg (84.83) vs. c+gcn+sg (83.84)} and \textsc{gcn+mg (84.40) vs. gcn+sg (80.50)}. 
The above results sufficiently establish that using segregated smaller sub-graphs as opposed to a single graph is more useful in learning richer vertex representations for relation extraction.

\subsection{Effect of Sentence Span}

To further assess the contribution of multiple sub-graphs over a single graph, we compare their performances using sentences with different lengths. 
For this purpose, we divide the SemEval test set into three groups (Table \ref{table:span_details}, $\mu=3, \sigma=9$) based on the distance between $e_1$ and $e_2$: 
(1) short spans $( k \leq \mu - \sigma)$; (2) medium spans $(\mu - \sigma < k < \mu + \sigma)$; and (3) long spans $(k \geq \mu + \sigma)$, where $\mu$ is the average number of tokens, and $\sigma$ is the standard deviation  over different lengths of tokens ($k$) between $e_1$ and $e_2$.

\begin{table}[t]
\begin{center}
{\small
\begin{tabular}{c c c  c}
\toprule
Short & Medium & Long & Total \\
\midrule
365 & 1966 & 386 & 2717 \\
13.50 (\%) & 72.30 (\%) & 14.20 (\%) \\
\bottomrule
\end{tabular}
}
\end{center}
\caption{\label{table:span_details} Total number of spans of different lengths and their percentage shares.}
\vspace{-5mm}
\end{table}

The best performing models using single graph (\textsc{c+gat+sg}) and multiple graphs (\textsc{c+gat+mg}) were examined on sentences in the above three categories as shown in Table \ref{table:multi_graph_contribution}. 
Interestingly, as seen in Table \ref{table:multi_graph_contribution}, different models using multiple graphs achieve a significantly higher performance than models using a single graph on sentences in the short span category. 
While \textsc{c+gat+mg} model without edge features, using a single graph achieves an F1-score of 73.09, the same model using multiple sub-graphs achieves an F1-score of 80.07. 
Although short span sentences form about 13.50\% of total sentences in the test set (Table \ref{table:span_details}), a significant improvement in the performance of models using multiple graphs on short spans contributes in achieving a higher score in the overall performance. 
The performance of models using multiple sub-graphs is also equally higher on long span sentences in comparison to models using a single graph. 
The ability of different models using multiple graphs to achieve a higher performance even in without edge features clearly shows that using multiple graphs with graph-based models is a useful method for relation extraction.

\begin{table}[t]
\begin{center}
{\small
\begin{tabular}{l c c  c}
\toprule
Model & P (\%) & R (\%) & F (\%) \\
\hline
  \multicolumn{4}{c}{\textsc{Models using Single Graph}} \\
 \hline 
 & \multicolumn{3}{c}{\textsc{short spans}} \\
 \midrule
\textsc{c+gat+sg} & 73.07 & 74.85 & 73.09 \\
\textsc{c+gat+sg+ctef} & 75.28 & 76.86 & 75.48 \\
\textsc{c+gat+sg+dref} & 73.93 & 73.97 & 73.34 \\
\textsc{c+gat+sg+ctef+dref} & 75.97 & 76.69 & 75.56 \\
\midrule
 & \multicolumn{3}{c}{\textsc{medium spans}} \\
\midrule
\textsc{c+gat+sg} & 84.01 & 86.95 & 85.42 \\
\textsc{c+gat+sg+ctef} & 84.75 & 84.33 & 84.43 \\
\textsc{c+gat+sg+dref} & 83.50 & 85.14 & 84.24 \\
\textsc{c+gat+sg+ctef+dref} & 84.22 & 84.42 & 84.26 \\
\midrule
 & \multicolumn{3}{c}{\textsc{long spans}} \\
\midrule
\textsc{c+gat+sg} & 72.12 & 75.62 & 73.69 \\
\textsc{c+gat+sg+ctef} & 74.47 & 73.70 & 73.85 \\
\textsc{c+gat+sg+dref} & 76.06 & 78.56 & 76.98 \\
\textsc{c+gat+sg+ctef+dref} & 76.91 & 75.43 & 75.80 \\ \\
\midrule
 \multicolumn{4}{c}{\textsc{Models using Multiple Graphs}} \\
\hline
 & \multicolumn{3}{c}{\textsc{short spans}} \\
 \midrule
\textsc{c+gat+mg} & 83.18 & 78.90 & 80.07 \\
\textsc{c+gat+mg+ctef} & 83.15 & 78.32 & 79.92\\
\textsc{c+gat+mg+dref} & 81.17 & 80.39 & 80.55\\ 
\textsc{c+gat+mg+ctef+dref} & 86.03 & 82.45 & 83.93 \\
\midrule
 & \multicolumn{3}{c}{\textsc{medium spans}} \\
\midrule
\textsc{c+gat+mg} & 87.83 & 84.94 & 86.27 \\
\textsc{c+gat+mg+ctef} & 87.77 & 86.24 & 86.94 \\
\textsc{c+gat+mg+dref} & 87.53 & 88.22 & 87.82\\ \textsc{c+gat+mg+ctef+dref} & 85.65 & 87.22 & 86.36 \\
\midrule
 & \multicolumn{3}{c}{\textsc{long spans}} \\
\midrule
\textsc{c+gat+mg} & 83.86 & 77.75 & 80.04 \\
\textsc{c+gat+mg+ctef} & 82.52 & 78.54 & 79.70 \\
\textsc{c+gat+mg+dref} & 77.37 & 80.25 & 78.43 \\
\textsc{c+gat+mg+ctef+dref} & 77.65 & 77.57 & 77.37 \\
\bottomrule
\end{tabular}
}
\end{center}
\caption{\label{table:multi_graph_contribution} Performance of models across short, medium and long spans in SemEval test set. P: Precision, R: Recall, F: F1-score}
\vspace{-5mm}
\end{table}

\subsection{Influence of Graph Size}

To evaluate the impact of graph size, we compare the performance of \textsc{c-gat-mg-dref} model under different graph sizes in Table \ref{table:graph_size}. Therein, \textsc{c+gat+mg+dref} indicates a graph limited to vertices in SDP; \textsc{c+gat+mg+dref\_1} is where first-order child vertices connected to the vertices in SDP are added in the graph; and \textsc{c+gat+mg+dref\_2} is where second and higher order child vertices associated with the vertices in SDP are included in the graph. 
As seen from Table \ref{table:graph_size}, the performance of \textsc{c+gat+mg+dref} model decreases with the graph size, indicating that distantly-connected vertices do not provide information relevant to the target relation.

\begin{table}[t]
\begin{center}
{\small
\begin{tabular}{l c c  c}
\toprule
Model & P & R & F \\
\midrule
\textsc{c+gat+mg+dref} & 86.03 & 86.66 & \textbf{86.30} \\
\textsc{c+gat+mg+dref}\_1 & 85.97 & 85.26 & 85.56 \\
\textsc{c+gat+mg+dref}\_2 & 84.81 & 86.04 & 85.36 \\
\bottomrule
\end{tabular}
}
\end{center}
\caption{\label{table:graph_size} Performance of \textsc{c+gat+dref+mg} on different graph sizes. P: Precision; R: Recall; F: F1-score}
\end{table}

\subsection{Comparisons against State-of-the-art}

The proposed \textsc{c+gat+mg+dref} model achieves the best F1-score of 86.3 against the SoTA graph-based models for relation extraction as shown in Table \ref{table:sota_performance}. The \textsc{c+gat+mg+dref} model scores higher than the C-AGGCN \cite{guo2019attention} model that selectively attends to relevant sub-structures by considering the full dependency graph. 
Moreover, the GCN-based model (\textsc{c-gcn-mg-dref}) using multiple graphs achieves a higher F1-score of 85.9 compared to the C-GCN model \cite{zhang2018graph}, which scores an F1-score of 84.8 using a pruned dependency tree along with the GCN model.

\begin{table}[t]
\begin{center}
\begin{tabular}{l c}
\toprule
Model Details & F1-Score\\
\midrule
SVM \cite{rink2010utd} & 82.2 \\
RNN \cite{socher2012semantic} & 77.6 \\
MVRNN \cite{socher2012semantic} & 82.4 \\
FCM \cite{yu2014factor} & 83.0 \\
CR-CNN \cite{santos2015classifying} & 84.1 \\
SDP-LSTM \cite{xu2015classifying} & 83.7 \\
DepNN \cite{liu2015dependency} & 83.6 \\
PA-LSTM \cite{zhang2017position} & 82.7\\
C-GCN \cite{zhang2018graph} & 84.8 \\
SPTree \cite{miwa2016end} & 85.5 \\
C-AGGCN \cite{guo2019attention} & 85.7 \\
\midrule
\multicolumn{2}{c}{\textsc{our models}} \\
\midrule
\textbf{\textsc{c-gcn-mg-dref}} & \textbf{85.9} \\
\textbf{\textsc{c-gat-mg-dref}} & \textbf{86.3} \\
\bottomrule
\end{tabular}
\end{center}
\caption{\label{table:sota_performance} Performance of the proposed model against state-of-the-art graph-based models for relation extraction.}
\vspace{-5mm}
\end{table}

\section{Conclusion}

We proposed a contextualised graph attention network using edge features and operating on multiple sub-graphs for relation classification. The proposed sub-graph partition method learns rich vertex representations for relation classification. 
We proposed two sets of edge features using dependency relations and connecting entity types and showed that by combining such edge features with GAT we establish a new state-of-the-art on the SemEval relation classification benchmark dataset.
The experimental results showed that using multiple sub-graphs is better than using a single graph with graphical networks such as GCNs and GATs.

\bibliography{acl2020}
\bibliographystyle{acl_natbib}

\end{document}